\documentclass[letterpaper, 10 pt, conference]{ieeeconf}  

\IEEEoverridecommandlockouts                              

\usepackage{xcolor}

\usepackage[T1]{fontenc}  
\usepackage{amsmath}      
\usepackage{bm}
\usepackage{amsfonts}
\usepackage{graphicx}
\usepackage{stackengine}
\usepackage{float}
\usepackage{caption}
\usepackage{subcaption}
\usepackage{makecell}

\makeatletter
\renewcommand*\env@matrix[1][*\c@MaxMatrixCols c]{%
  \hskip -\arraycolsep
  \let\@ifnextchar\new@ifnextchar
  \array{#1}}
\makeatother

\title{\LARGE \bf
A Unified Model with Inertia Shaping for Highly Dynamic Jumps of Legged Robots }

\author{Ke Wang$^{1}$, Guiyang Xin$^{2}$, Songyan Xin$^{2}$, Michael Mistry$^{2}$, Sethu Vijayakumar$^{2}$ and Petar Kormushev$^{1}$
\thanks{$^{1}$Robot Intelligence Lab, Dyson School of Design Engineering,
Imperial College London, UK. Email: {\tt\small k.wang17@imperial.ac.uk}}%
\thanks{$^{2}$School of Informatics, University of Edinburgh,
Edinburgh, UK.}%
}

\begin{document}

\maketitle
\thispagestyle{empty}
\pagestyle{empty}

\begin{abstract}
To achieve highly dynamic jumps of legged robots, it is essential to control the rotational dynamics of the robot. In this paper, we aim to improve the jumping performance by proposing a unified model for planning highly dynamic jumps that can approximately model the centroidal inertia. This model abstracts the robot as a single rigid body for the base and point masses for the legs. The model is called the Lump Leg Single Rigid Body Model (LL-SRBM) and can be used to plan motions for both bipedal and quadrupedal robots. By taking the effects of leg dynamics into account, LL-SRBM provides a computationally efficient way for the motion planner to change the centroidal inertia of the robot with various leg configurations. Concurrently, we propose a novel contact detection method by using the norm of the average spatial velocity. After the contact is detected, the controller is switched to force control to achieve a soft landing. Twisting jump and forward jump experiments on the bipedal robot SLIDER and quadrupedal robot ANYmal demonstrate the improved jump performance by actively changing the centroidal inertia. These experiments also show the generalization and the robustness of the integrated planning and control framework.


\end{abstract}

\section{INTRODUCTION}

Early studies on highly dynamic motion such as jumping and running in legged robots are largely influenced by the heuristic control implemented on Raibert’s hoppers \cite{raibert1986legged}. The hopper is able to achieve the dynamic behaviors through simple composition of a set of simple controllers that control hopping height, speed, and posture separately. This is possible due to their prismatic leg design which differs from most humanoid robots with human-like morphology. 
Despite its success, heuristic control is quite limited since it needs large amount of work on parameter tuning. More recently, model based approaches are becoming more and more popular. The spring loaded inverted pendulum (SLIP) model is a well recognized template model for running and jump. 
An approximated SLIP model has been used to represent the translational motion, robust running and jump are planned with model predictive control \cite{mordatch2010robust}.
3D-SLIP model has been used to generate high speed running motion \cite{wensing2013high} and long jump \cite{wensing2014development}. However, the SLIP model only considers the point mass dynamics and the angular momentum is ignored. It is therefore difficult to consider motions involving body rotation, such as a twisting jump. 

 \begin{figure}
  \centering
    \begin{subfigure}{0.8\columnwidth}
      \includegraphics[width=0.9\linewidth]{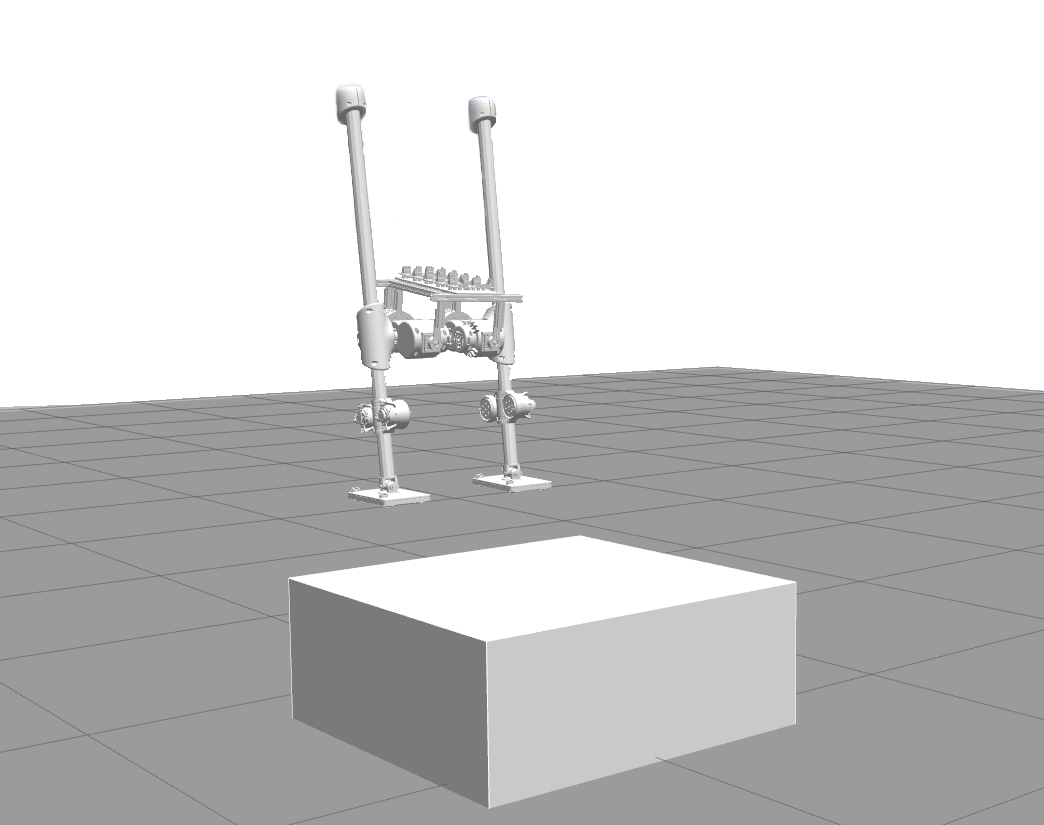} 
      \label{fig:intro_1}
    \end{subfigure} 
    \begin{subfigure}{0.8\columnwidth}
      \vspace{0.3cm}
      \includegraphics[width=0.9\linewidth]{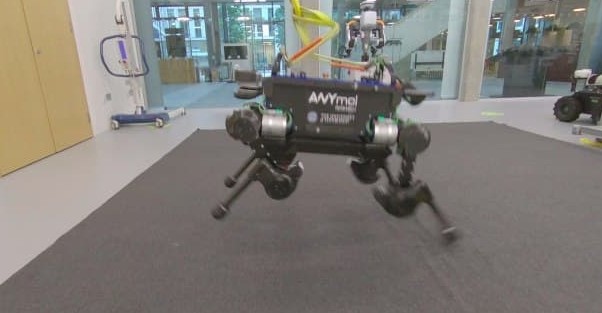}
      \label{fig:intro_2}
    \end{subfigure}\hfil
  \caption{Highly dynamic jump motions can be generated for both bipedal and quadrupedal robots using the LL-SRBM. Top: SLIDER robot. Bottom: ANYmal robot.}
\vspace{-5mm}
\end{figure}

Compared to humanoid robots, quadrupedal robots are often built with lower degree of freedom (DoF) legs and point feet. This equips them with lightweight legs which are often ignored during the motion planning phase or the control phase without introducing significant modeling error. 
With the assumption of massless legs, dynamic gaits including trotting, pronking, bounding and pacing have been demonstrated on Mini Cheetah through online convex model predictive control \cite{di2018dynamic,kim2019highly,katz2019mini}. The convex formulation is made possible due to a small angle assumption for roll and pitch angle of the Single Rigid Body Model (SRBM) and predefined footstep locations. Unfortunately, these assumptions do not hold for generating versatile and highly dynamic motions.
Recently, a more versatile and acrobatic motion generation framework, the kino-dynamic planner, is proposed in \cite{chignoli2021mit}. However, the kino-dynamic planner is complex and computationally inefficient. Therefore, a good trade-off between model complexity and computation efficiency is needed. 

At the same time, the robustness of the controller is important for guaranteeing the successful execution of the dynamic motion. The robot experiences a large impact when landing; these large contact forces and fast changing velocities can easily make the robot unstable. Roboticists have made various efforts to reduce the effect of the impact. One direction is changing the reference trajectory in real time. \cite{corberes:hal-03034022} shows a quadrupedal robot walking with 50 Hz re-planning with a slightly simplified SRBM. But no results have been shown with more dynamic motions such as jumping. In \cite{halm2019modeling, 9341246} the impacts are explicitly modelled, however the simplified model and large uncertainties of state estimation make the control more difficult. \cite{yang2021impact} presents a feedback controller in an impact invariant space. Although this approach is not affected by the rapid change of velocities, it can not reduce the effects caused by the impact. To reduce the effect of the impact, good contact detection with a soft landing is a promising direction. External force sensors are commonly used to monitor contacts for robot arms \cite{haddadin2017robot}. But in practice force signals are too noisy for legged robots. The average spatial velocity, first introduced by \cite{4209816}, is a synthetic representation of all link velocities and an indicator for the kinetic energy of the robot. We will rely on the average spatial velocity for contact detection.


\subsection{Contributions}
In this paper, we aim to improve the performance of highly dynamic jumps with the LL-SRBM and the robustness of the tracking controller with contact detection. Additionally, we show that the proposed planning and control methods can be applied to both bipedal and quadrupedal robots. The twisting jump and forward jump are demonstrated in simulation and real robot experiments. Here, we would like to highlight the contributions of the paper:

\begin{itemize}
        \item {A general framework that can generate highly dynamic jumping motions for both bipedal and quadrupedal robots. Each layer of the framework is generalized for both types of robots, including the lump leg concept. }
        \item {The Lumped Leg Single Rigid Body Model (LL-SRBM) provides a computationally efficient way to optimize the shape of the robot's inertia while planning the jump motion. By taking the motion of legs into consideration, the jump motion generated by the LL-SRBM achieves better performance (e.g. faster twisting jump, more stable forward jump etc.) than the motion generated by using the SRBM. }
        \item {Safe landing is ensured by using a novel contact detection method and switching to force control in landing phase. We propose to use the norm of the average spatial velocity to detect contacts. By taking account of all link states, the average spatial velocity is more robust to noises and general enough for detecting impulses from all directions. After the contact is detected, contact force tracking control is switched on for a period to achieve soft landing. We cannot use force control for the whole jump process since the LL-SRBM is still an approximated model for planning.}

\end{itemize}


\section{SYSTEM OVERVIEW}

\begin{figure}[t]
\centering
\includegraphics[width=0.9\columnwidth]{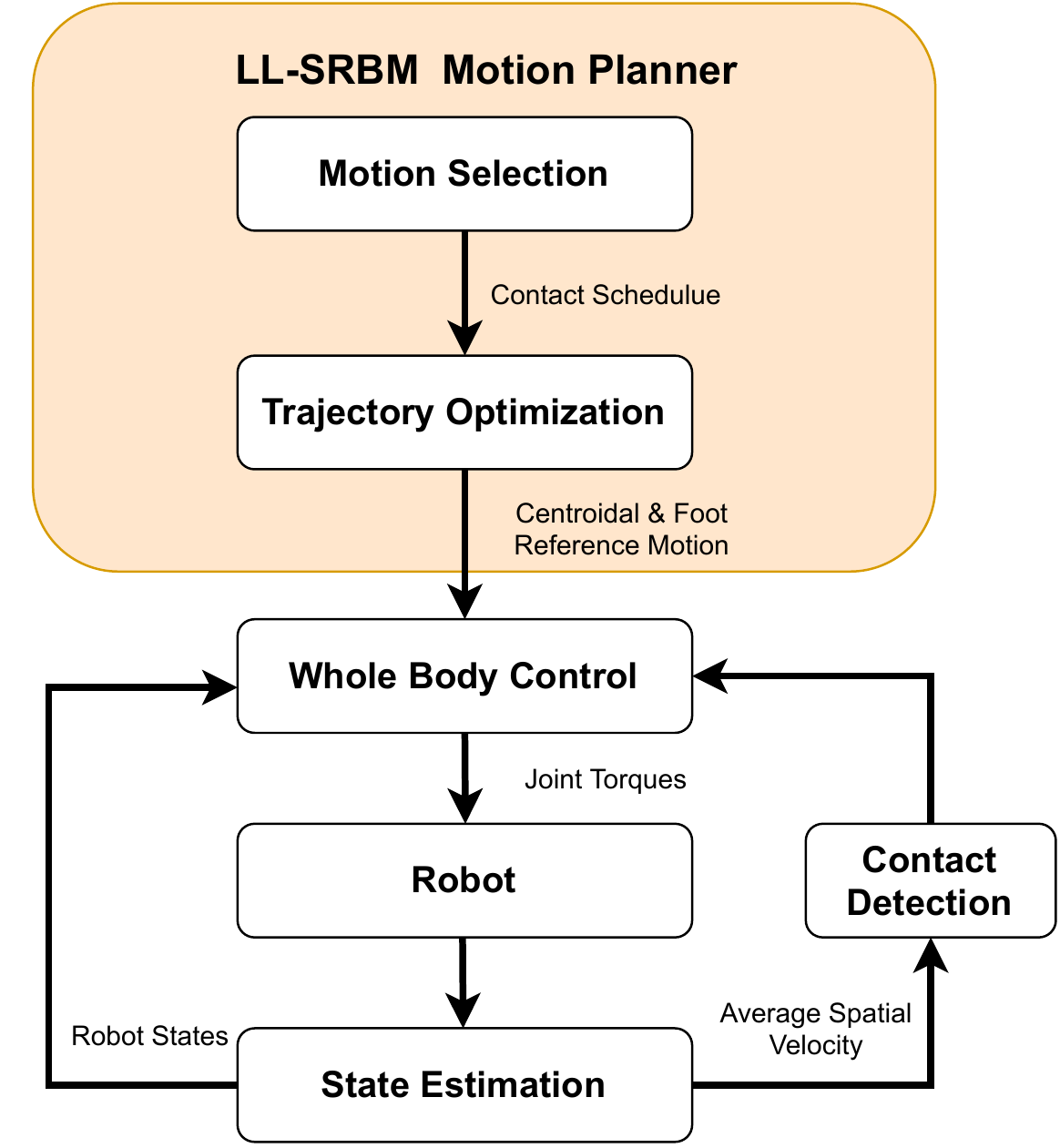}
\caption{The planning and control framework.}
\label{fig:control_hierarchy}
\vspace{-5mm}
\end{figure}

The highly dynamic jumping behaviour is achieved via a hierarchical structure which includes the motion planner and the whole body controller as shown in Fig. \ref{fig:control_hierarchy}.  
The high-level user selects the motion (e.g. twisting jump, forward jump, etc.) and inputs to the motion planner with initial and final states and contact positions for non-flight phases. The trajectory optimization with LL-SRBM outputs the reference centroidal and foot trajectories to the whole body controller. The whole body controller computes optimal joint torques for the robot to track these task space targets while considering all physical constraints. The contact detection runs after the robot takes off and monitors the average spatial velocity. Once the contact is detected, the whole body controller switches to force control for a short time to achieve a soft landing. After landing, trajectory tracking is reinstated to achieve the desired steady finishing pose.

\section{The Lump Leg Single Rigid Body Model}
\begin{figure}[t]
\centering
\begin{subfigure}{0.49\columnwidth}
  \includegraphics[width=1.1\linewidth]{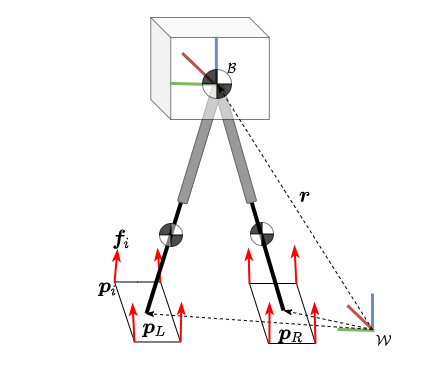} 
  \caption{}
  \label{fig:model_schema_1}
\end{subfigure} 
\begin{subfigure}{0.49\columnwidth}
  \includegraphics[width=1.1\linewidth]{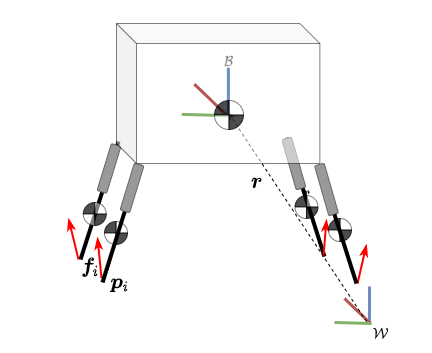}
  \caption{}
  \label{fig:model_schema_2}
\end{subfigure}\hfil
\caption{The LL-SRBM is a general model for bipedal and quadrupedal robots, it takes the leg dynamics into consideration. It also unifies planar, line and point feet with the distributed Ground Reaction Force representation. (a) shows LL-SRBM applied to a bipedal robot with planar feet. The ground reaction forces are located at four corners of the foot with arbitrary directions. (b) shows LL-SRBM applied to a quadrupedal robot with point feet. For both figures $\mathcal{W}$ and $\mathcal{B}$ stand for the world frame and the body frame.}
\label{fig:LLSRBM}
\vspace{-5mm}
\end{figure}

\begin{figure}[t]
\label{fig:foot_schematic}
\centering
\begin{subfigure}{0.48\columnwidth}
  \includegraphics[width=1.0\linewidth]{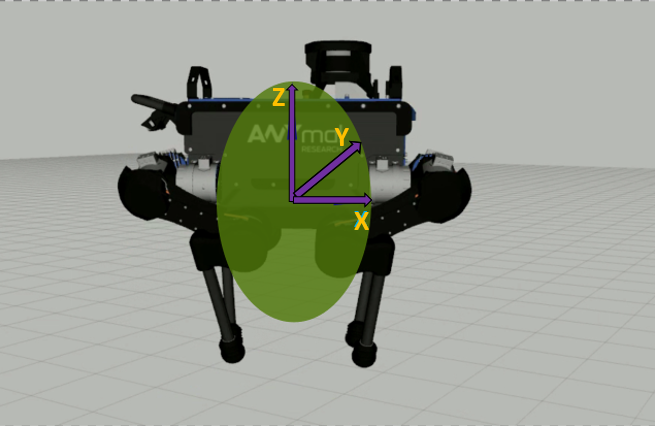} 
  \caption{}
  \label{fig:model_ill_1}
\end{subfigure} 
\begin{subfigure}{0.48\columnwidth}
  \includegraphics[width=1.0\linewidth]{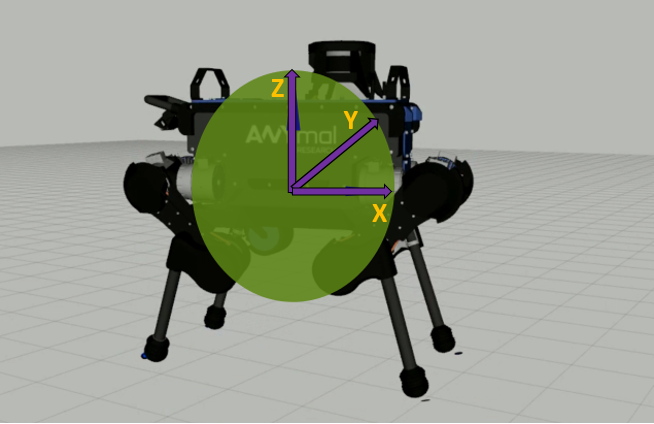}
  \caption{}
  \label{fig:model_ill_2}
\end{subfigure}\hfil
\caption{The shape of the centroidal inertia of the robot is affected by the leg configuration, which can be modelled by LL-SRBM. (a) and (b) show two different leg configurations when ANYmal is performing a twisting jump. Compared with the centroidal inertia in (b), the centroidal inertia in (a) has smaller values on \textit{y} and \textit{z} axes.}
\label{fig:inertia_shape_illustrate}
\vspace{-5mm}
\end{figure}

We are interested in generating high fidelity dynamic jump motions while keeping the model simple enough. The novel aspect of LL-SRBM model is that it abstracts the robot as a single rigid body for the base and point masses for the legs, as illustrated in Fig. \ref{fig:LLSRBM}. The point mass of the leg is located at the CoM of the leg at the default configuration and the distance between the point mass and the leg contact location is proportional to the leg length. Compared to SRBM, the LL-SRBM models the leg dynamics as well, which is important for computing the centroidal inertia, as show in Fig. \ref{fig:inertia_shape_illustrate}.  The full Newton-Euler dynamics of the LL-SRBM with multiple contacts with the environment can be written as:
\begin{equation}
    \label{eq1}
 \begin{bmatrix}
        \dot{\bm{H}} = \sum^{n}_{i=1}\bm{f}_{i} + m\bm{g} \\
\dot{\bm{L}} = \sum^{n}_{i=1}(\bm{p}_{i} - \bm{r})\times \bm{f}_{i}
    \end{bmatrix}
\end{equation}
where $\bm{H}$ and $\bm L$ are the linear and angular momentum of the model, $\bm f_{i}$ is the $i$th ground reaction force acting on $\bm p_{i}$, $n$ forces are assumed. $\bm r$ is the position of the CoM, $m$ and $\bm g$ are the mass of the model and the gravity vector.
The linear momentum is directly related to the translational velocity of the CoM: 
\begin{equation}
    \label{eq2}
    \dot{\bm{r}} = \bm{H}/m \\
\end{equation}

For LL-SRBM, the CoM is computed as 
\begin{equation}
    \label{eq2-1}
    \bm{r} = \frac{\bm{r}_B\cdot m_B+\sum_i{\bm{r}_l^i \cdot m_l^i}}{m_B+\sum_i{m_l^i}}
\end{equation}
where $\bm{r}_B$ and $m_B$ are CoM and mass of the body, $\bm{r}_l^i$ and $m_l^i$ are the position and mass of the $i$th leg. 

Angular momentum of the LL-SRBM, $\bm L$, can be expressed as:
\begin{equation}
    \label{eq3}
    \bm L = \bm{I}_{W}\bm{\omega} 
\end{equation}
where $\bm I_{W}$ represents the centroidal inertia of the model in the world frame and $\bm{\omega}$ is the angular velocity of the body. The centroidal inertia matrix, also called the centroidal composite rigid body inertia (CCRBI) matrix in \cite{4209816} can be computed as:
\begin{equation}
    \label{eq4}
    \bm{I}_W =\bm{R}_B\bm{I}_G\bm{R}_B^{T}
    =\bm{R}(\bm{q}_{B})\bm{I}_G\bm{R}^{T}(\bm{q}_{B})
\end{equation}
where $\bm{R}(\bm{q}_{B})$ is a transformation from local to world frame as given in the Appendix. $\bm{I}_G \in \mathbb{R}^{6 \times 6}$ is the centroidal inertia matrix at CoM frame and can be computed as:
\begin{equation}
    \label{eq5}
    \bm{I}_G = \bm{I}_B + \sum_i{\bm{I}_l^i} = \bm{I}_B + \sum_i{m_i(\bm{r}-\bm{p}_i+\Delta\bm{p}_i)^2}
\end{equation}
where $\bm{I}_B$ is the inertia of the single rigid body projected to the CoM, and $\Delta\bm{p}_i$ is the offset between the leg's contact location and the position of leg mass, the second term represents the inertia of legs projected to CoM. 

In \cite{di2018dynamic} the Euler angles are used to represent the orientation, however this approach suffers from the problem of gimbal lock, so we prefer quaternion representation for the orientation. The dynamics of the quaternion $\bm{q}_{B}$  can be expressed as:
\begin{equation}
    \label{eq6}
    \dot{\bm{q}}_{B} = \frac{1}{2}\bm{q}_{B} \circ \bm{\omega} = \bm{Q}(\bm{q}_{B})\bm{\omega}
\end{equation}
where $\circ$ represents the quaternion product and $\bm{Q}(\bm{q}_{B}) \in \mathbb{R}^{4 \times 3}$ is the corresponding matrix representation. 
By associating equation (\ref{eq3}), (\ref{eq4}), (\ref{eq6}), we can get the dynamics of orientation represented by quaternion:
\begin{equation}
    \label{eq7}
    \dot{\bm{q}}_{B} = \frac{1}{2}\bm{Q}(\bm{q}_{B})\bm I_{W}^{-1}(\bm{q}_{B})\bm{L}
\end{equation}

So the complete dynamics of the LL-SRBM can be summarized as:
\begin{equation}
    \label{eq8}
    \left. \begin{bmatrix}
        \dot{\bm{r}} = \bm{H}/m \\
        \dot{\bm{q}}_{B} = \frac{1}{2}\bm{Q}(\bm{q}_{B})\bm I_{W}^{-1}(\bm{q}_{B})\bm{L}\\
        \dot{\bm{H}} = \sum^{n}_{i=1}\bm{f}_{i} + m\bm{g} \\
        \dot{\bm{L}} = \sum^{n}_{i=1}(\bm{p}_{i} - r)\times \bm{f}_{i} \\
    \end{bmatrix}\right. 
\end{equation}
with
\begin{equation}
    \bm{r} = \frac{\bm{r}_B \cdot m_B+\sum_i{\bm{r}_l^i \cdot m_l^i}}{m_B+\sum_i{m_l^i}} 
      \tag{\ref{eq2-1}}
\end{equation}
\begin{equation}
    \bm{I}_W =\bm{R}(\bm{q}_{B})(\bm{I}_B + \sum_i{m_i(\bm{r}-\bm{p}_i+\Delta\bm{p}_i)^2})\bm{R}^{T}(\bm{q}_{B})
\end{equation}
    
\section{TRAJECTORY OPTIMIZATION}
This section describes the mathematical formulation of trajectory optimization and how we formulate trajectory optimization into a Nonlinear Programming (NLP) problem.

\subsection{Problem Formulation}

The trajectory optimization \cite{kelly2017introduction} is a powerful tool to find optimal trajectories for complex tasks that contain nonlinear dynamics and involves multiple phases. In this paper, we employ the \textit{multiple shooting method} to transcribe our trajectory optimization problem into a NLP formulation. In the formulation, multiple phases are considered since we are focusing on versatile dynamic jumps that contains multiple pre-defined contact phases: \textit{takeoff}, \textit{flight phase} and \textit{post-landing phase}.

 
 The system has been discretised into $N$ segments and therefore $N+1$ knots exists. Since $m$ phases are assumed, each phase gets $N/m$ segments.  
 Because timing is important for generating highly dynamic motions, we do not fix the time for each phase $\Delta T$. However we keep the same time interval $dt$ between knots inside each phase to reduce the computation cost. 
 For knot $k$, the state $\boldsymbol{x}[k]$ and control $\boldsymbol{u}[k]$ are defined as:
\begin{align*}
    \boldsymbol{x}[k] =& [ 
    \boldsymbol{r}[k], 
    \boldsymbol{q}_B[k], 
    \boldsymbol{H}[k], 
    \boldsymbol{L}[k]]
    \\
    \boldsymbol{u}[k] =& [
    \boldsymbol{f}_{1}[k], ..., 
    \boldsymbol{f}_{N_l}[k], 
    \boldsymbol{p}_{1}[k], ...,
    \boldsymbol{p}_{N_l}[k]
    ] \\
    \boldsymbol{\xi}[k] =& [\boldsymbol{x}[k], \boldsymbol{u}[k], dt[k]]
\end{align*}
where $dt[k]$ is the time interval from knot $k$ to $k+1$, $N_l$ is the leg number, $\boldsymbol{f}_{1}[k], ..., \boldsymbol{f}_{N_l}[k]$ are the ground reaction forces acting on the feet. For bipedal robots with planar feet, the left foot and right foot collect four corresponding corner forces:  
$\boldsymbol{f}_{L}[k]=[\boldsymbol{f}_{L_1}[k], \boldsymbol{f}_{L_2}[k], \boldsymbol{f}_{L_3}[k], \boldsymbol{f}_{L_4}[k]]$,
$\boldsymbol{f}_{R}[k]=[\boldsymbol{f}_{R_1}[k], \boldsymbol{f}_{R_2}[k], \boldsymbol{f}_{R_3}[k], \boldsymbol{f}_{R_4}[k]]$ as shown in Fig. \ref{fig:model_schema_1}.
$\boldsymbol{p}_{L}[k]$ and $\boldsymbol{p}_{R}[k]$ are the center position of the left foot and right foot respectively. For quadrupedal robots, $    \boldsymbol{u}[k] = [
    \boldsymbol{f}_{1}[k], 
    \boldsymbol{f}_{2}[k], 
    \boldsymbol{f}_{3}[k], 
    \boldsymbol{f}_{4}[k], 
    \boldsymbol{p}_{1}[k],
    \boldsymbol{p}_{2}[k],
    \boldsymbol{p}_{3}[k],
    \boldsymbol{p}_{4}[k]
    ]$, as represented by Fig. \ref{fig:model_schema_2}.


Given the open parameters $\boldsymbol{\xi}[k]$, the complete trajectory optimization problem can be formulated as follows:

\begin{align}
\label{eq:complete formulation}
    \min_{\boldsymbol{\xi}}  \sum_{k=0}^{N}l(\boldsymbol{\xi}[k]) + \phi(\boldsymbol{\xi}[N])  \tag*{(cost function)} \\
    \textrm{s.t.} \quad  {\boldsymbol{x}}[k+1] = f(\boldsymbol{x}[k], \boldsymbol{u}[k])  \tag*{(LL-SRBM dynamics)} \\
    \boldsymbol{x}[0] = \boldsymbol{x}_{ini}^{} \tag*{(initial states)} 
    \\
    \boldsymbol{x}[N] = \boldsymbol{x}_{fin}^{} \tag*{(final states)} 
    \\
    t_{min} \leq  dt[k] \leq t_{max} \tag*{(time limits)} \\
    \lVert \boldsymbol{q}_B[k] \rVert = 1 \tag*{(quaternion norm)}
\end{align}
\begin{align}
    l_{min} \leq \lVert\boldsymbol{r}[k] - \boldsymbol{p}_{i}[k] \rVert \leq l_{max} \quad (i\in \{1, ..., N_l\})
    \tag*{(kinematics limits)} 
\end{align}
 \quad if foot in {contact}:
\begin{align}
    {0} \leq {f}_{i}^z \leq {f_{max}^z} \quad (i\in\{1,...,N_l\})  \tag*{(ground force limit)} 
    \\
    \boldsymbol{P} \bm{f}_{i}  \leq \boldsymbol{0} \quad (i\in\{1,...,N_l\}) \tag*{(friction cone)} 
    \\
    \boldsymbol{p}_{i}[k] = \boldsymbol{p}_{i}^* \quad (i\in\{1,...,N_l\}) \tag*{(given contacts)} 
\end{align}
\quad if foot in the {air}:
\begin{align}
        \bm{f}_{i} =  \boldsymbol{0} \quad (i\in\{1,...,N_l\})  \tag*{(no contact force)}
\end{align}

\subsection{Cost Function}
The items inside the cost function can be categorized into 4 groups: \textit{control inputs smoothness, energy consumption, time penalization} and \textit{final state target}.

\textit{Control Inputs Smoothness}: To generate a smooth motion, the difference of ground reaction forces and foot movements between adjacent nodes are minimized:
\begin{equation}
    \sum_{k=0}^{N}\sum_{i=0}^{N_l}\big(
    \dot{\boldsymbol{p}}_{i}^2[k] 
    + \dot{\boldsymbol{f}}_{i}^2[k] \big) 
\end{equation}

\textit{Energy Consumption}: We minimize squared momentum to achieve minimal energy consumption:
\begin{equation}
    \sum_{k=0}^{N} (\boldsymbol{H}^2[k] + \boldsymbol{L}^2[k])
\end{equation}

\textit{Time Penalization}: We want the optimization to generate a minimum-time trajectory to finish a task while respecting the constraints. Also, this cost term encourages the robot to change the shape of the centroidal inertia in the \textit{flight phase} to achieve a fast motion:

\begin{equation}
    \sum_{i=0}^{N}(dt^2[i])
\end{equation}

\textit{Final State Target}: Though final state targets are formulated as constraints, we find that putting final orientation target into cost function can speed up the solving process. This is especially the case in tasks requiring large change of orientation, like twisting jump. So we put the final orientation targets into the cost function:
\begin{equation}
    (\boldsymbol{q}_B[N]-\boldsymbol{q}_{B}^{fin})^2 + (\dot{\boldsymbol{q}}_B[N]-\dot{\boldsymbol{q}}_{B}^{fin})^2
\end{equation}
\subsection{Kinematics Constraint}
The kinematics constraint bounds the distance between CoM and the feet to be within $[l_{min},\, l_{max}]$, 
\begin{align}
   l_{min} \leq \lVert\boldsymbol{r}[k] - \boldsymbol{p}_{i}[k] \rVert \leq l_{max} \quad (i\in \{1,...,N_l\})
\end{align}
For the bipedal robot SLIDER, $l_{min}=$ 0.35~m and $l_{max}=$ 0.75~m. For the quadrupedal robot ANYmal,  $l_{min}=$ 0.31~m and $l_{max}=$ 0.6~m.
\subsection{Contact Constraint}
We have pre-defined the contact sequence as \textit{takeoff phase}, \textit{flight phase} and \textit{landing phase}. We pre-define the start and target position of the foot, namely:
\begin{align}
        \boldsymbol{p}_{i}[k] = \boldsymbol{p}_{i}^{ini} 
        \quad (i\in\{1,...,N_l\})
         \tag*{(takeoff phase)} 
        \\
        \boldsymbol{p}_{i}[k] = \boldsymbol{p}_{i}^{fin} 
        \quad (i\in\{1,...,N_l\}) \tag*{(post-landing phase)} 
\end{align}

\subsection{Ground Reaction Force Constraint}
In \cite{chignoli2021mit} the actuator limits are introduced in the planner by approximating the configuration-dependent reaction forces. Although joint angles of the robot are not included in LL-SRBM, we can still approximate the ground reaction force limit using the default configuration by:
\begin{equation}
    \bm{f}_{max} = \bm{J}^{\mathrm{T}}(\bm{q}_0)\bm{\tau}_{max}
\end{equation}
where $\bm{J}$ is the jacobian matrix, $\bm{q}_0$ represents the joint coordinates at the default configuration, $\bm{\tau}_{max}$ is the maximum joint torque vector. 
In \textit{takeoff phase} and \textit{landing phase} both feet need to be in contact with the ground, so all the ground reaction forces would be 0 or pushing against the ground, therefore the ground force limit is:
\begin{align}
{0} \leq {f}_{i}^z \leq {f^z_{max}} \quad (i\in\{1,...,N_l\}) 
\end{align}

We also ensure no slippage of foot contact points. The tangential forces are constrained to remain inside the Coulomb friction cone defined by the friction coefficient $\mu$. We approximate the friction cone by the friction pyramid, which is a common approach to make the constraints linear and speed up the computation. The friction cone constraint is given by:
\begin{align}
    -\mu f_{i}^{z} \leq f_{i}^{x} \leq \mu f_{i}^{z}\\
    -\mu f_{i}^{z} \leq f_{i}^{y} \leq \mu f_{i}^{z}
\end{align}
where $f_{i}^{x}$, $f_{i}^{y}$, $f_{i}^{z}$ are components of the ground reaction force $\bm{f}_{i} \, (i\in\{1,...,N_l\})$.

If all feet is in the air, which is the case of \textit{flight phase}, we enforce all ground reaction force to be exactly 0:
\begin{align}
        \bm{f}_{i} =  \boldsymbol{0} \quad (i\in\{1,...,N_l\})
\end{align}

\section{WHOLE-BODY CONTROL and CONTACT DETECTION}
The whole-body controller takes responsibility of computing the joint torques to achieve the desired motions while respecting a set of constraints. In the paper, the tasks of interest are the CoM position, the pelvis orientation, the angular momentum of the robot, the foot positions and orientations. Each task is comprised of a desired acceleration as a feed-forward term and a state feedback term to stabilize the trajectory. Generally, the task for the linear motion can be expressed as:
\begin{equation*}
    \bm{J}_{\mathrm{T}}\ddot{\bm{q}} =  \ddot{\bm{x}}^{\mathrm{cmd}} - \dot{\bm{J}}_{\mathrm{T}}\dot{\bm{q}},
\end{equation*}
\begin{equation*}
    \ddot{\bm{x}}^{\mathrm{cmd}} = \ddot{\bm{x}}^{\mathrm{des}} + \bm{K}_{\mathrm{P}}^{\mathrm{pos}}(\bm{x}^{\mathrm{des}} - \bm{x}) + \bm{K}_{\mathrm{D}}^{\mathrm{pos}}(\dot{\bm{x}}^{\mathrm{des}} - \dot{\bm{x}}),
\end{equation*}
where $\bm{J}_{\mathrm{T}}$ is the translational Jacobian for the task, $\bm{x}$ is the actual position of the link, and the superscript $\mathrm{des}$ indicates the desired motion. 

For the task of angular motion, the command can be formulated as:
\begin{equation*}
    \bm{J}_{\mathrm{R}}\ddot{\bm{q}} =  \dot{\bm{\omega}}^{\mathrm{cmd}} - \dot{\bm{J}}_{\mathrm{R}}\dot{\bm{q}},
\end{equation*}
\begin{equation*}
    \dot{\bm{\omega}}^{\mathrm{cmd}} = \dot{{\bm \omega}}^{\mathrm{des}} + \bm{K}_{\mathrm{P}}^{\mathrm{ang}}(\mathrm{AngleAxis}(\bm{R}^{\mathrm{des}}\bm{R}^{\mathrm{T}})) +  \bm{K}_{\mathrm{D}}^{\mathrm{ang}}(\bm{\omega}^{\mathrm{des}} - \bm{\omega}),
\end{equation*}
where $\bm{J}_{\mathrm{R}}$ is the rotational Jacobian for the task, $\bm{R}$ and $\bm{R}^{\mathrm{des}}$ denote the actual and desired orientation of the pelvis link respectively, 
$\mathrm{AngleAxis}()$ maps a rotation matrix to the corresponding axis-angle representation, 
$\bm {\omega} \in \mathbb{R}^3$ is the angular velocity of the link.

For the CoM task and angular momentum task, the centroidal momentum matrix \cite{orin2013centroidal} is used as the task jacobian. 
For balancing or walking, angular momentum task are often defined as a damping task that damps out excess angular momentum. However for highly dynamic motion such as twisting jump, angular momentum varies a lot during the process and it plays a vital role to achieve the jump motion. In this case, the reference angular momentum is needed and it comes from our motion planner. Since the single rigid body model is used in the motion planner, its orientation and associated angular momentum are both well defined.

\subsection{QP formulation}
Inspired by \cite{Herzog_2015}, the full dynamics can be decomposed into the underactuated part and actuated part:
\begin{equation*}
    \begin{bmatrix}
    \bm{M}_f \\ \bm{M}_a 
    \end{bmatrix} \bm{\ddot{q}} +
    \begin{bmatrix}
        \bm{H}_f \\ \bm{H}_a
    \end{bmatrix} = 
    \begin{bmatrix}
        \bm{0} \\ \bm{S}_a
    \end{bmatrix} \bm\tau +
    \begin{bmatrix}
        \bm{J}^{\mathrm{T}}_f \\ \bm{J}^{\mathrm{T}}_a
    \end{bmatrix} \bm{f},
\end{equation*}
where $\bm{M}$, $\bm{H}$, $\bm{S}_a$, $\bm {\tau}$, $\bm{J}$ and $\bm{f}$ are the mass matrix, Coriolis force matrix and gravitation force vector, the actuator selection matrix, joint torques vector, the stacked contact Jocabian and reaction force vector. The subscript, $f$ and $a$, indicates the floating part and actuated part respectively. The weighted sum formulation is applied, in which one QP problem is solved at each control loop.
The formulation of the QP problem can be written as 
\begin{align}
\min_{\ddot{\bm{q}},\, \bm{f}} \quad & \frac{1}{2}\| \bm{A}\ddot{\bm{q}} + \dot{\bm{A}}\dot{\bm{q}} - \bm{B}^{\mathrm{cmd}} \|_{\bm{W}_1}^2 + \frac{1}{2}\| \bm{f}-\bm{f}_{\mathrm{des}} \|^2_{\bm{W}_2}\\
\textrm{s.t.} \quad & \bm{M}_f\bm{\ddot{q}} - \bm{J}^{\mathrm{T}}_f\bm{f} = - \bm{H}_f \tag*{(floating base dynamics) }\\
& \bm{P}\bm{f} \leq \bm{0}\tag*{(friction cone)}\\
& \bm{S}_a^{-1}(\bm{M}_a \ddot{\bm{q}} + \bm{H}_a - \bm{J}^{\mathrm{T}}_a\bm{f}) \in [\bm{\tau}_{min},\, \bm\tau_{max}]\tag*{(input limits)}
\end{align}

where $\bm{A}$ is a stack of the Jacobian matrices for the tasks of interest, $\bm{B}^{\mathrm{cmd}}$ is a stack of the commanded accelerations and $\bm{W}_i \, (i = 1, \,2)$ are the weighting matrices, $\bm{P}$ denotes the linearized friction cone matrix. Similar to~\cite{apgar2018fast}\cite{kuindersma2016optimization} , the unilateral contact constraint is treated as a soft constraint by simply assigning a large weight on the desired zero acceleration. It is reported in \cite{feng2014optimization} that this gives a better stability.  

The output torque commands $\bm \tau$ at each control iteration is computed by
\begin{equation}
    \bm\tau = \bm{S}_a^{-1}(\bm{M}_a \bm{\ddot{q}} + \bm{H}_a - \bm{J}^{\mathrm{T}}_a\mathbf{f})
\end{equation}

Since the SRBM is an approximate model, we set $\bm{W}_2 = \bm{0}$ to track the desired trajectory in the jump motion, except the landing phase. In the landing phase, in order to achieve soft landing, we switch to tracking desired contact forces by increasing $\bm{W}_2$ while decreasing the weights for centroidal momentum control to be $\bm{0}$. After a short safe landing phase (around 0.2 second), we switch back to trajectory tracking control in order to achieve the desired steady pose. This strategy of switching to force control during landing is similar to \cite{chignoli2021mit}, which is particular critical for real robot experiments in the existing of trajectory tracking and state estimation errors. 

\subsection{Contact detection}
According to \cite{orin2013centroidal}, the spatial average velocity of a robot is defined as
\begin{equation*}
    \mathbf{v}_G=\begin{bmatrix}
        \bm{v}_{\mathrm{com}} \\
        \bm{\omega}_G
    \end{bmatrix} = \bm{I}_f^{-1}\bm{h}_G
\end{equation*}
where $\bm{h}_G \in \mathbb{R}^6$ is the centroidal momentum of the robot, $\bm{I}_f \in \mathbb{R}^{6 \times 6}$ is the centroidal inertia matrix of the robot, $\bm{v}_{\mathrm{com}}$ is the center of mass velocity, $\bm{\omega}_{G}$ denotes the average angular velocity of the robot. Please refer to \cite{orin2013centroidal} for the details of $\bm{h}_G$ and $\bm{I}_f$.

We use the change rate of $\| \mathbf{v}_G \|_2$ to judge the contact. During the flight phase, the gravity is the only external force that changes the state of the robot. The norm of $\bm{v}_{\mathrm{com}}$ will increase under the gravity while the robot is dropping down from the peak. When contacts happen, the velocities will decrease. In order to respond to all kinds of contact, we choose to use $\| \mathbf{v}_G \|_2$ to detect contacts which is a comprehensive metric including all the changes of velocities of each link of a robot. When the numerical derivatives of $\| \mathbf{v}_G \|_2$ in a sampling window are all less than a threshold, we tell the controller that the robot gets contact. In practice, the proposed method is more reliable than using force signals that are very noisy. 
\section{EXPERIMENT RESULTS}
\subsection{Robot Platforms}

SLIDER is a knee-less bipedal robot designed by the Robot Intelligence Lab at Imperial College London \cite{9341143,  wang2018clawar}. It is 1.2~m tall and has 10 actuated joints with a total weight of 16~Kg. Most of its weight is concentrated in the pelvis. The prismatic knee joint design is an unique feature of this robot that differentiates it from many other robots with anthropomorphic design.
Also the sliding joint has a relatively large range of motion. The overall light weight and large range of leg motion make the robot suitable for agile locomotion. 

The quadrupedal robot ANYmal is made by ANYbotics. It has 12 SEAs (Series Elastic Actuators) and weighs approximately 35~Kg. When ANYmal standing still in the default configuration, it is about 0.5~m tall.

\begin{figure}[t]
\begin{subfigure}{0.155\textwidth}
  \includegraphics[width=1.02\linewidth]{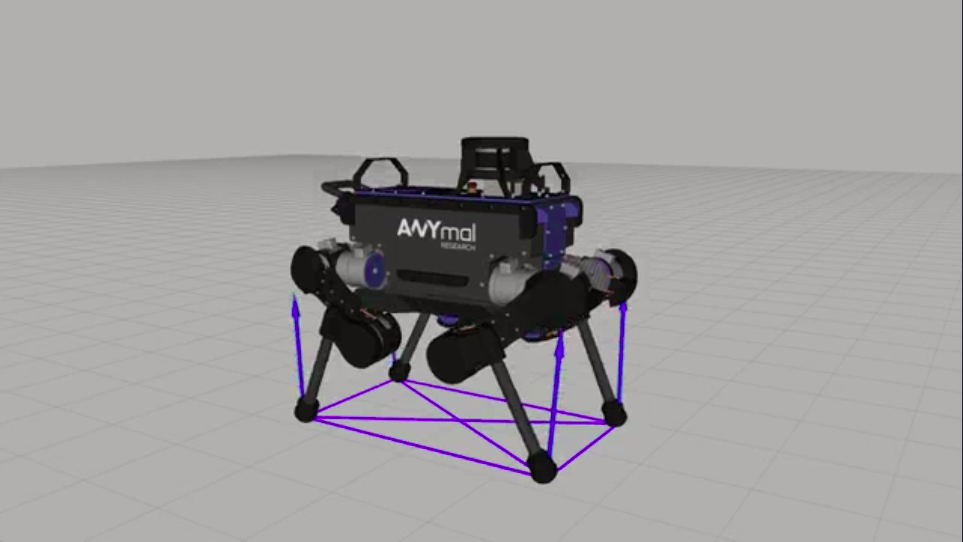} 
\end{subfigure} 
\begin{subfigure}{0.155\textwidth}
  \includegraphics[width=1.02\linewidth]{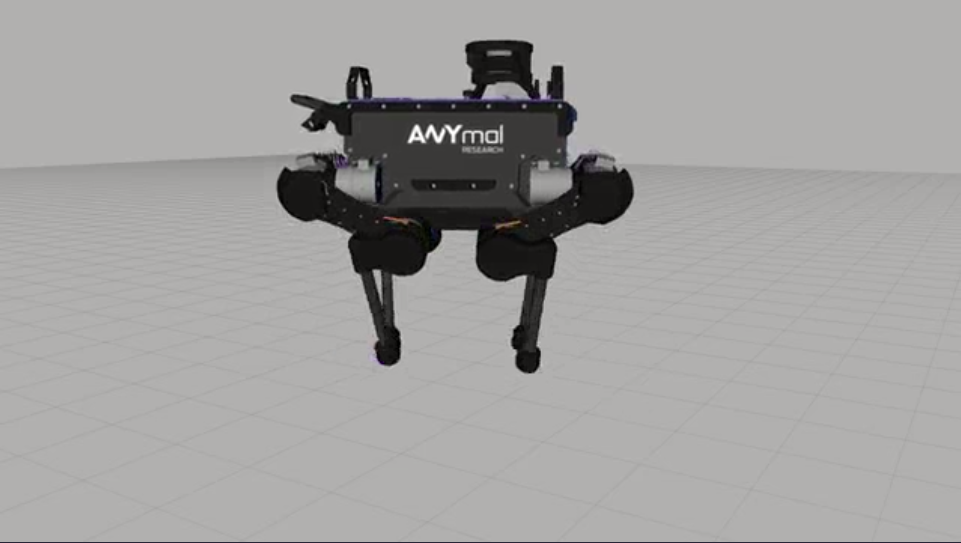}
\end{subfigure}\hfil
\begin{subfigure}{0.155\textwidth}
  \includegraphics[width=1.02\linewidth]{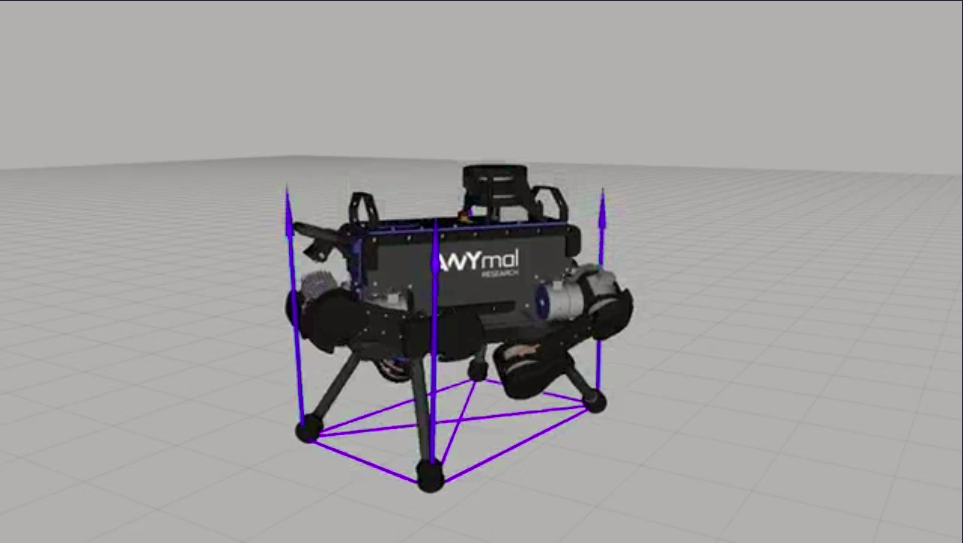}
\end{subfigure}\hfil

\medskip
\begin{subfigure}{0.155\textwidth}
  \includegraphics[width=1.02\linewidth]{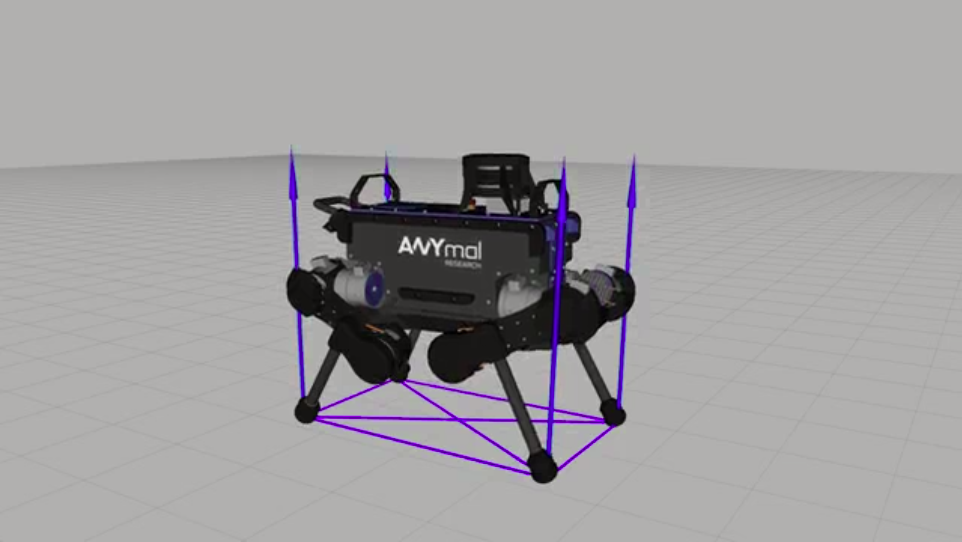}
\end{subfigure} 
\begin{subfigure}{0.155\textwidth}
  \includegraphics[width=1.02\linewidth]{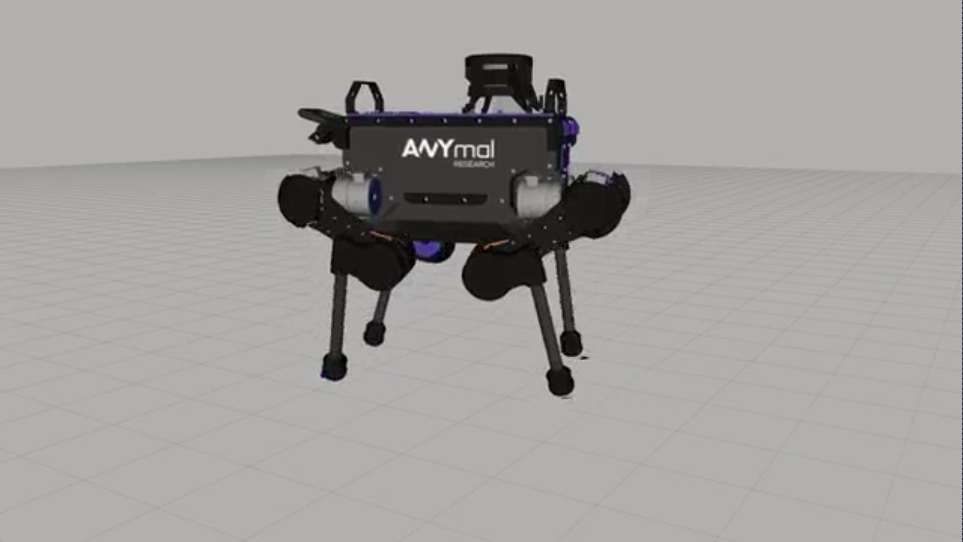}
\end{subfigure} \hfil
\begin{subfigure}{0.155\textwidth}
  \includegraphics[width=1.02\linewidth]{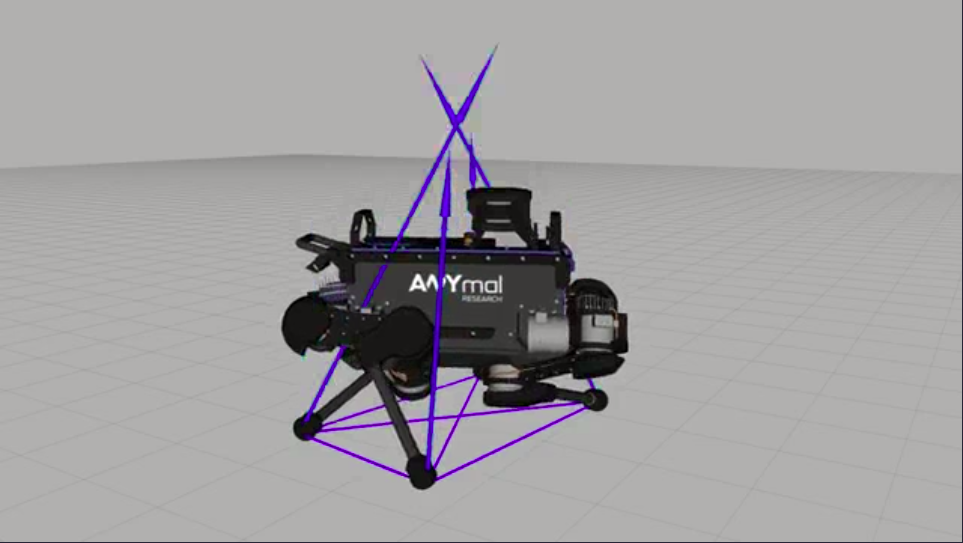}
\end{subfigure} \hfil
\caption{Snapshots of the ANYmal robot performing a $90^{\circ}$ twisting jump with inertia shaping and without inertia shaping. First row: twisting jump with inertia shaping. Second row: twisting jump without inertia shaping.}
\label{fig:jumpTwist_snapshot}
\end{figure}

\begin{figure}[t]
    \centering 
\begin{subfigure}{0.46\columnwidth}
  \includegraphics[width=1.1\linewidth]{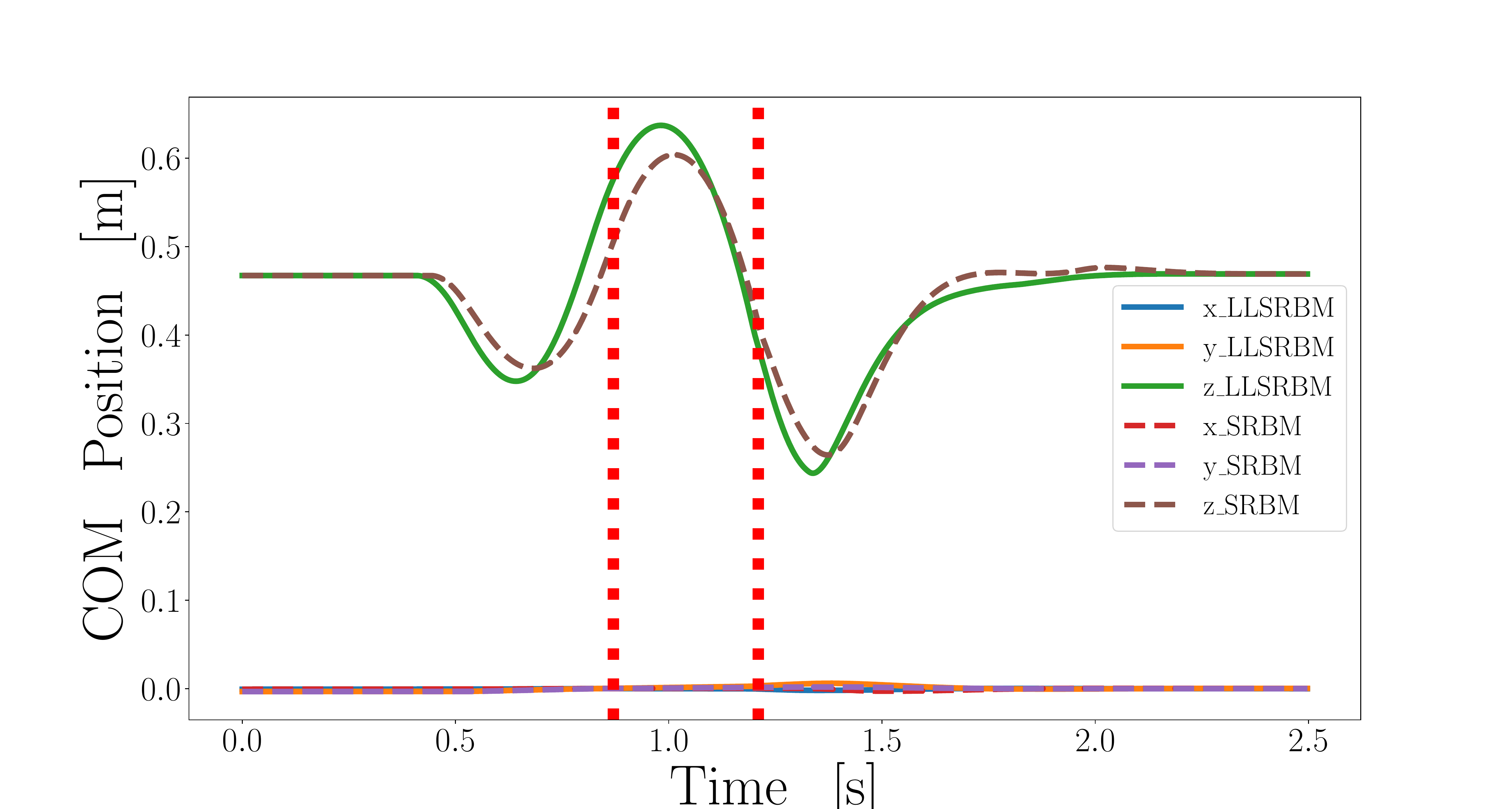} 
\end{subfigure} 
\begin{subfigure}{0.46\columnwidth}
  \includegraphics[width=1.1\linewidth]{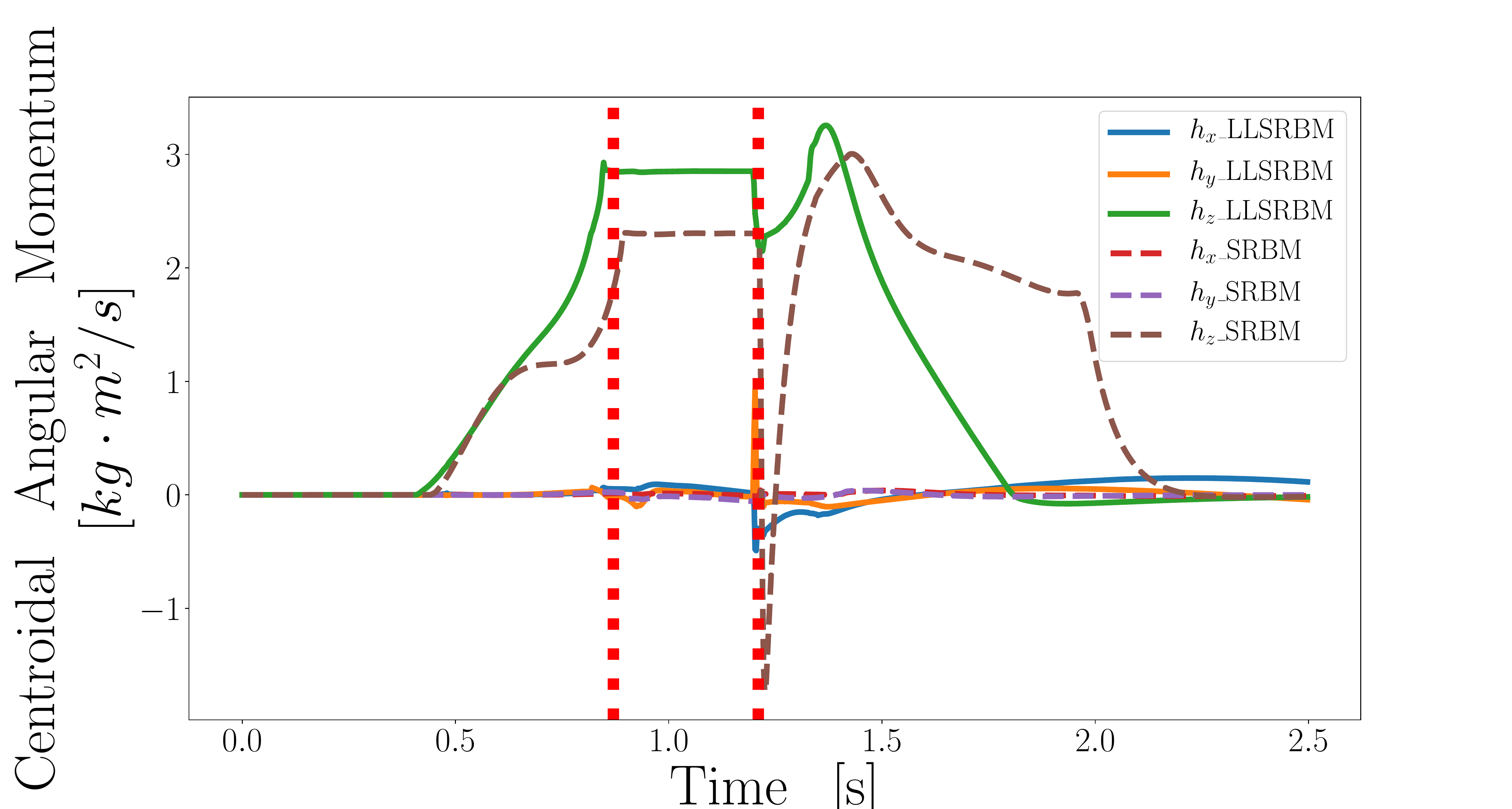}
\end{subfigure}

\begin{subfigure}{0.46\columnwidth}
  \includegraphics[width=1.1\linewidth]{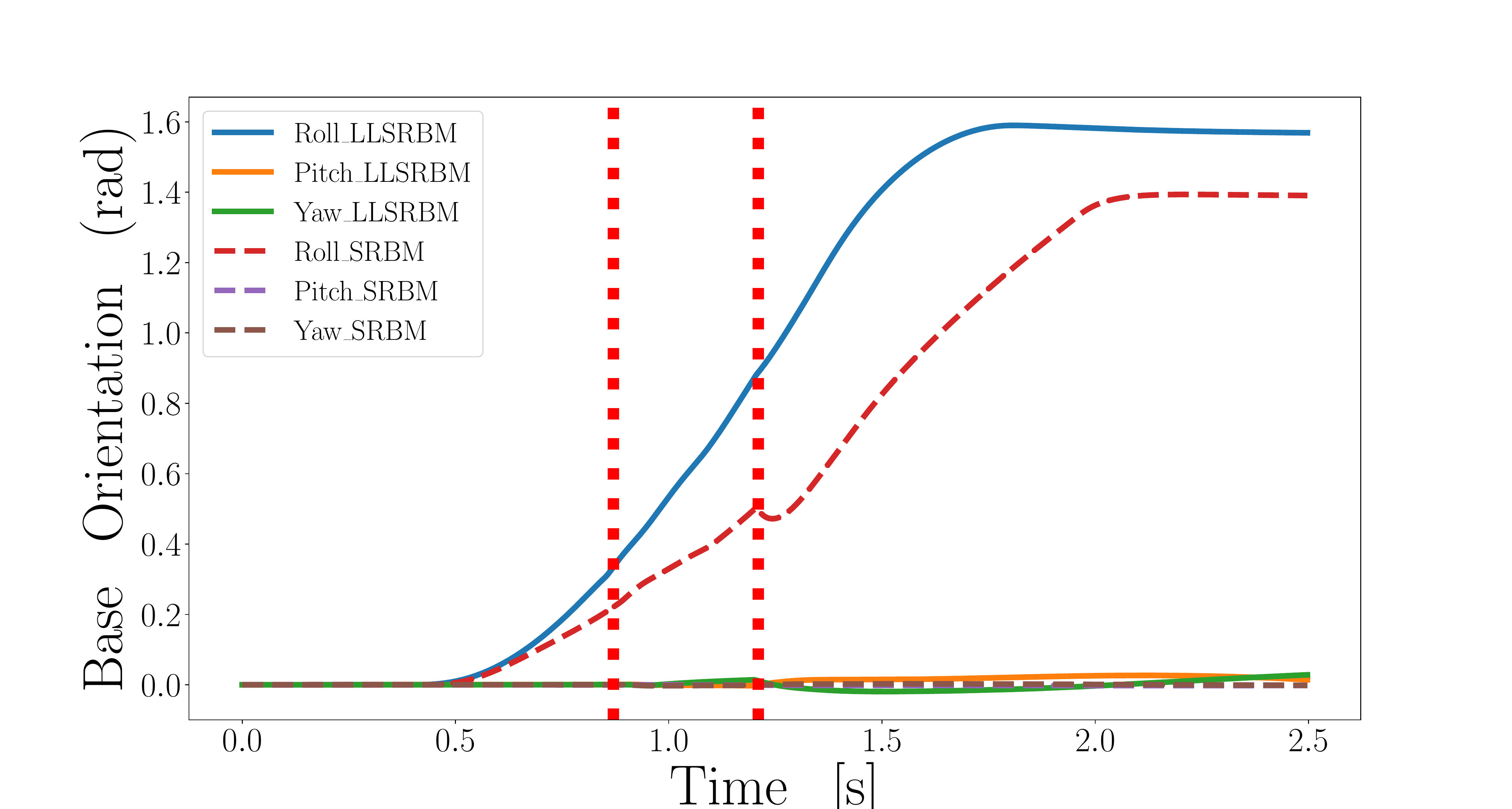}
\end{subfigure} 
\begin{subfigure}{0.46\columnwidth}
  \includegraphics[width=1.1\linewidth]{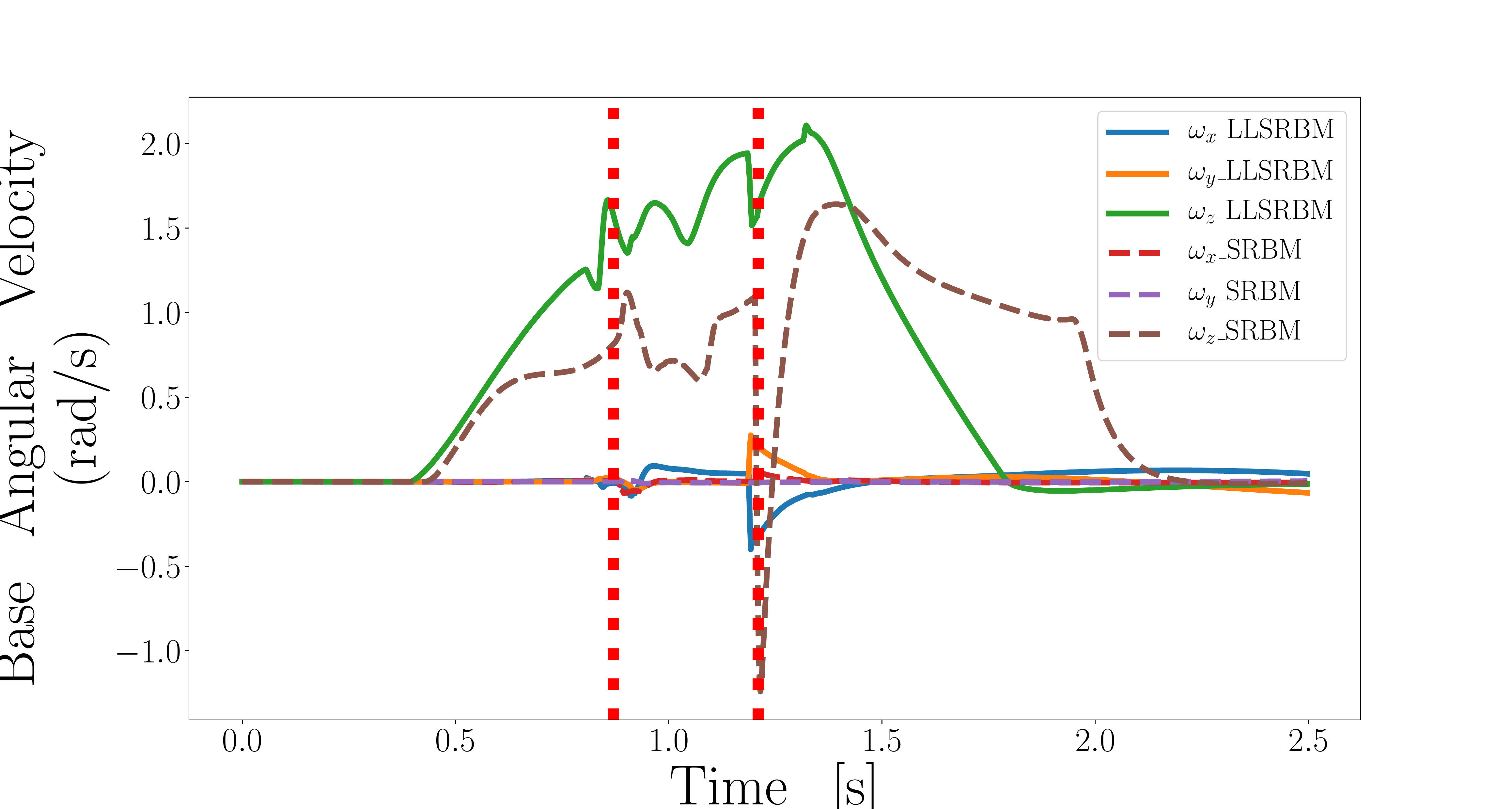}
\end{subfigure} \hfil
\caption{The measured trajectories of ANYmal $90^{\circ}$ twisting jump in simulation, with measured trajectories generated by LL-SRBM and SRBM repectively. In each plot the vertical dashed lines split the entire trajectory into three phases: \textit{takeoff}, \textit{flight} and \textit{landing}. The full lines show the measured trajectories with LL-SRBM and the dashed lines show measured trajectories with SRBM.}
\label{fig:jumpTwist_plot}
\vspace{-5mm}
\end{figure}
\subsection{Implementation}
The trajectory optimization framework is implemented in CasADi \cite{andersson2019casadi} with Python using the interior-point solver IPOPT \cite{wachter2006implementation}. We write the whole body controller with C++ using the Pinocchio library \cite{pinocchioweb} to compute full rigid body dynamics and qpOASES \cite{ferreau2014qpoases} to solve the QP problem. For SLIDER robot the whole body controller runs at 1k~Hz to track the desired trajectory. For ANYmal the whole-body controller is running at 400~Hz in simulation and real experiments as well. For both robots the simulation is done in Gazebo with ODE as the physics engine.

\subsection{Twisting Jump}
We experimented the twisting jump on both ANYmal and SLIDER robots and compared the performance between the trajectories generated by LL-SRBM and SRBM. In the trajectory optimization we simply provide a linear interpolation of the orientation as the initial guess. Also we assign a large weight to the \textit{final state target} and \textit{time penalization} in the cost function to encourage the solver to find a trajectory that reaches the target pose in a minimum time. 

As shown in Fig. \ref{fig:jumpTwist_snapshot}, the foot motion generated with LL-SRBM has a big difference compared to the one generated with SRBM. In the flight phase, the feet try to keep close to the central rotating axis with the LL-SRBM generated trajectories. The SRBM generated foot trajectories just do an interpolation from the start to the goal position. This interesting behaviour emerges as LL-SRBM takes leg dynamics into consideration in the planning and tries to maximize the angular velocity in flight phase by modifying the shape of inertia with the leg motion. Figure \ref{fig:jumpTwist_plot} shows the measured trajectories of the twisting jump experiments. It can be seen clearly that in flight phase the base angular velocity in experiments with LL-SRBM does not drop as much as the base angular velocity in experiments with SRBM. The robot reaches the goal orientation earlier with LL-SRBM. To accomplish the twisting jump for SLIDER, we have added two joints for the robot, one yaw joint per leg. SLIDER can perform a $90^{\circ}$ twisting jump and the motion can be seen in the accompany video. 

\begin{figure}[b]
\vspace{-5mm}
    \centering
    \includegraphics[width=0.9\linewidth]{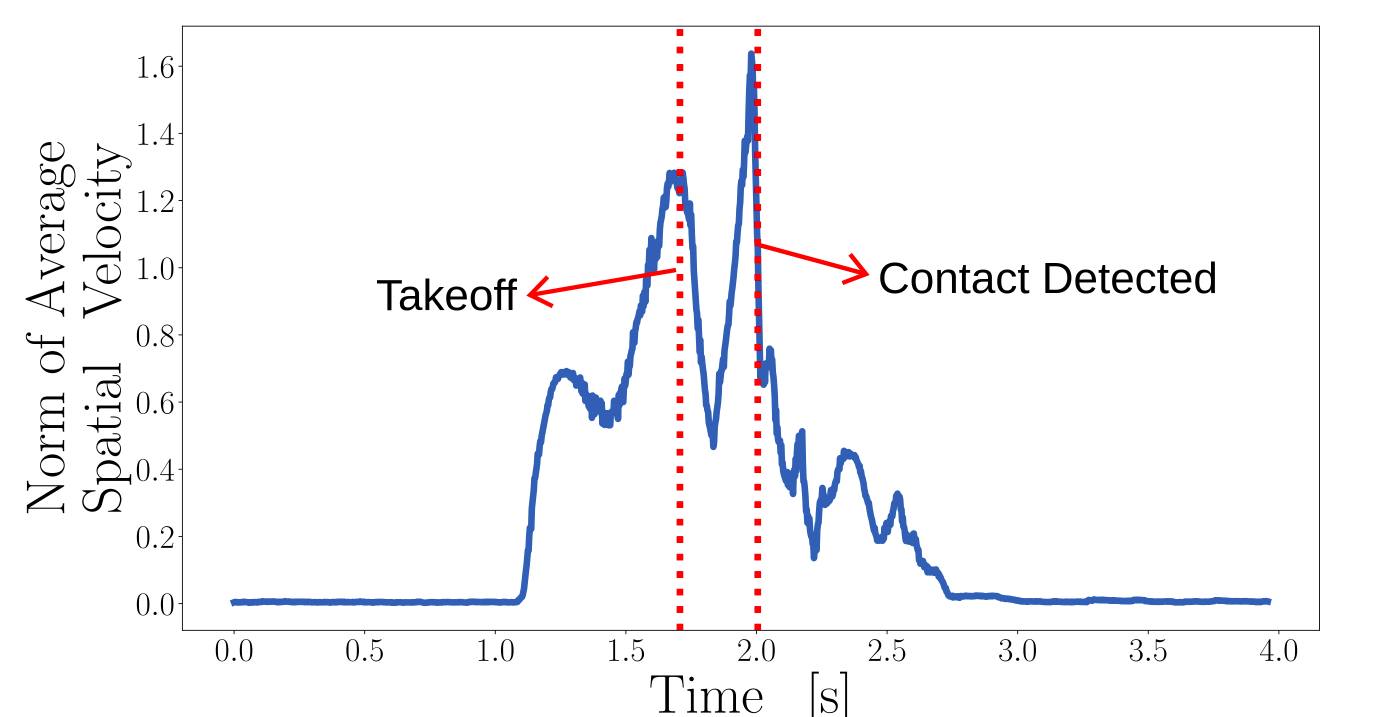}
    \caption{The norm of average spatial velocity when ANYmal performed a twisting jump. A wooden board was put under the robot while it was flying. A big drop of the value after getting contact when the robot was dropping down.}
    \label{fig:twist_average_spatial}
\end{figure}

\begin{figure*}[t]
\centering 
\begin{subfigure}{0.24\textwidth}
  \includegraphics[width=1.0\linewidth]{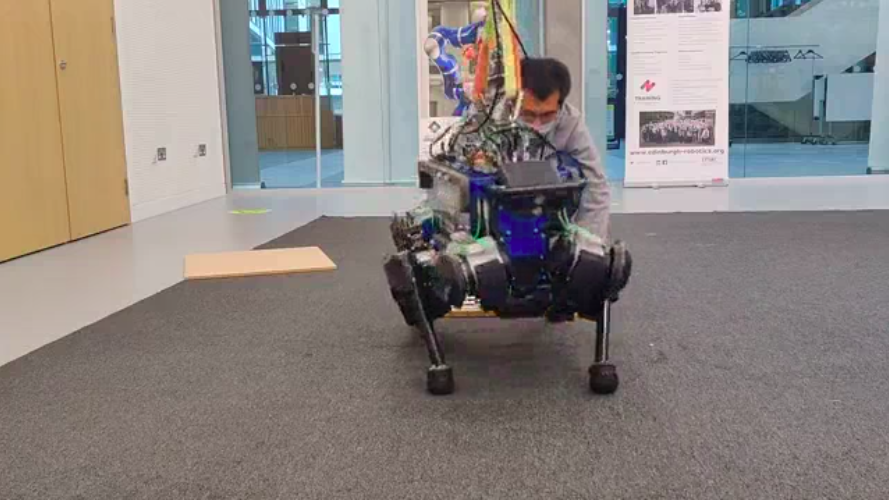}
  \caption{}
\end{subfigure}
  \hspace{0.01cm}
\begin{subfigure}{0.24\textwidth}
  \includegraphics[width=1.0\linewidth]{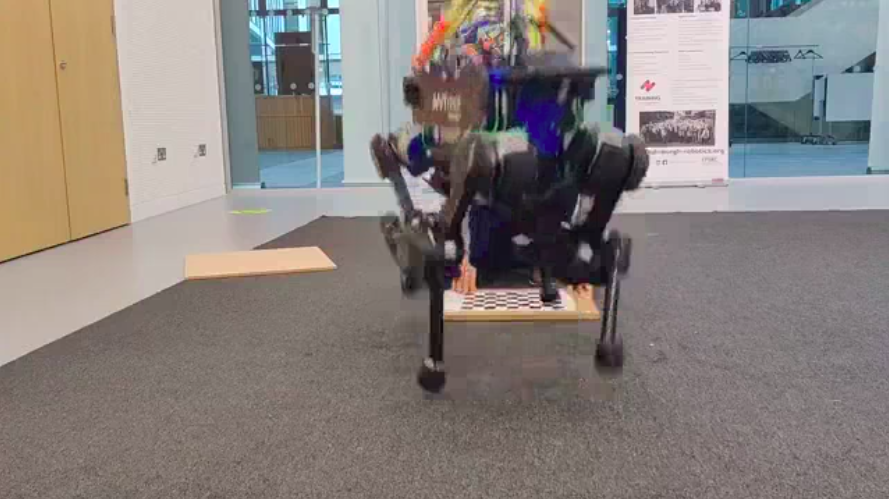}
  \caption{}
\end{subfigure} 
\begin{subfigure}{0.24\textwidth}
  \includegraphics[width=1.0\linewidth]{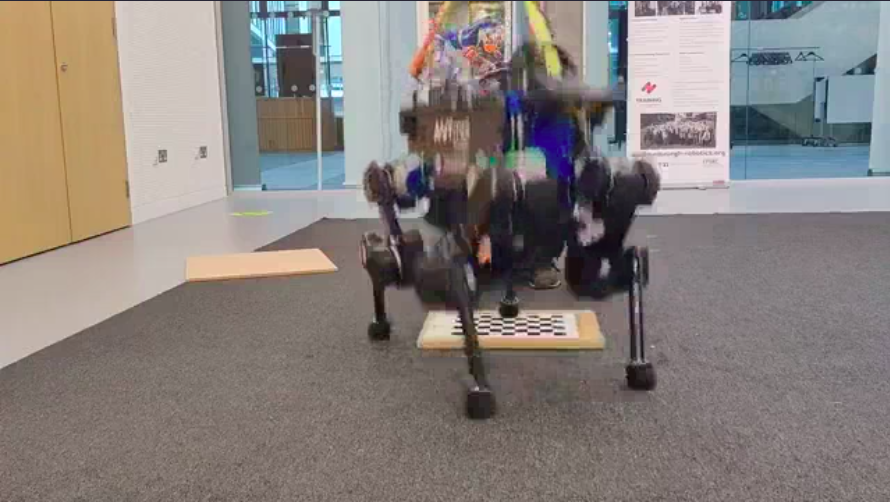}
  \caption{}
\end{subfigure}
  \hspace{0.01cm}
\begin{subfigure}{0.24\textwidth}
  \includegraphics[width=1.0\linewidth]{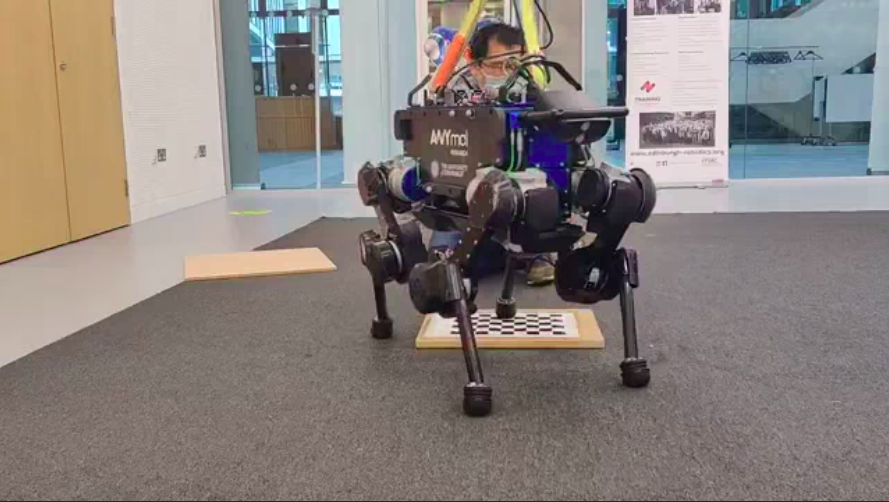}
  \caption{}
\end{subfigure} 
\caption{Snapshots of ANYmal performing a twisting jump with $30^{\circ}$. We also put a wooden board with a height of 25~mm under the robot when it was in the \textit{flight} phase to test the contact detection and the robustness of our controller. From left to right: $\text{(a)}$: takeoff phase, $\text{(b)}$: flight phase, $\text{(c)}$: contact is detected. $\text{(d)}$: the robot is stable after a soft landing.}
\label{fig:twistJump_real}
\vspace{-5mm}
\end{figure*}
In real experiments, we demonstrate the effectiveness of the proposed approaches by twisting jumps on unknown objects, as shown in Fig \ref{fig:twistJump_real}. A wooden board with a height of 25~mm was put under ANYmal when it was in the flight phase. Figure \ref{fig:twist_average_spatial} shows the norm of the average spatial velocity. Even though collected from real robot experiments, the norm of the average spatial velocity is smooth and can be used as a reliable judgement for contact detection. The robot successfully detected the contact event and switched to a soft landing.

\subsection{Forward Jump}
\begin{figure}[t]
\begin{subfigure}{0.155\textwidth}
  \includegraphics[width=1.02\linewidth]{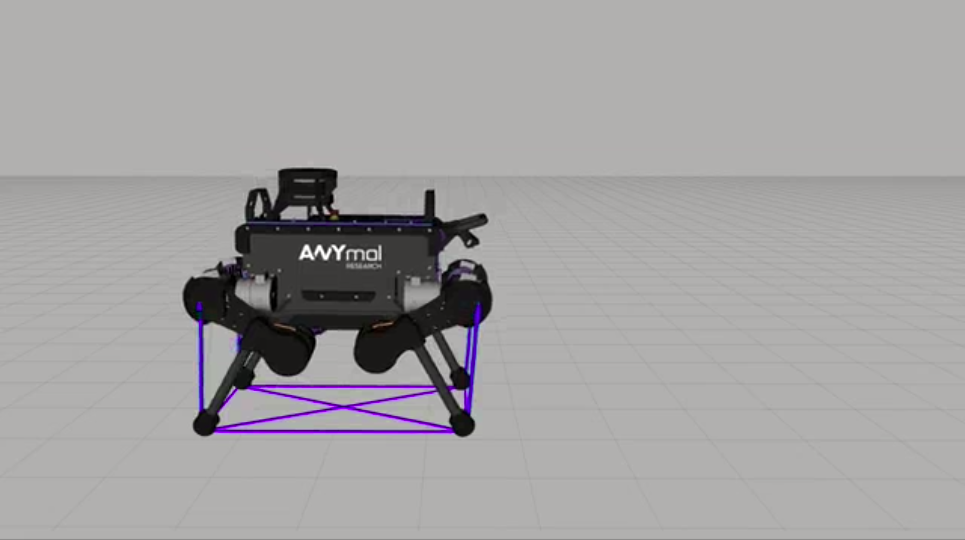} 
\end{subfigure} 
\begin{subfigure}{0.155\textwidth}
  \includegraphics[width=1.02\linewidth]{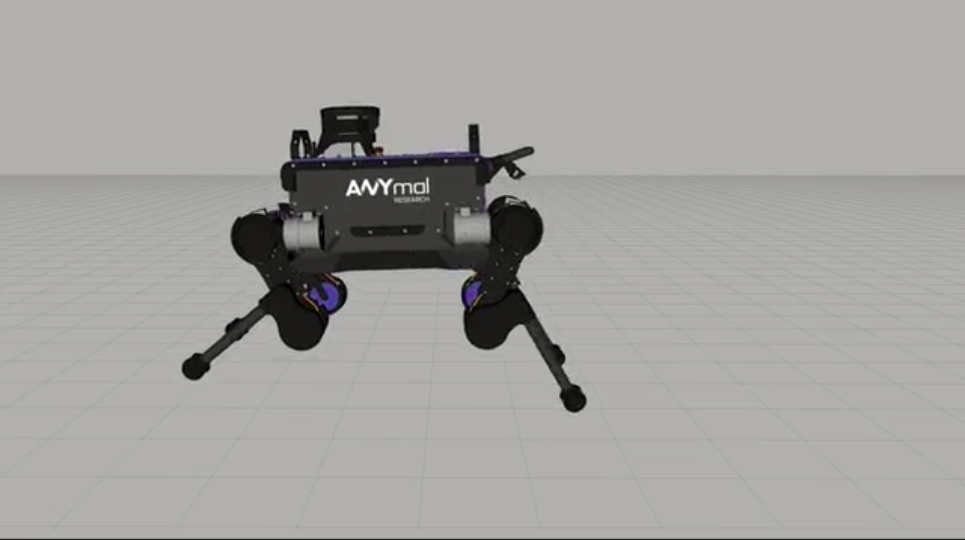}
\end{subfigure}\hfil
\begin{subfigure}{0.155\textwidth}
  \includegraphics[width=1.02\linewidth]{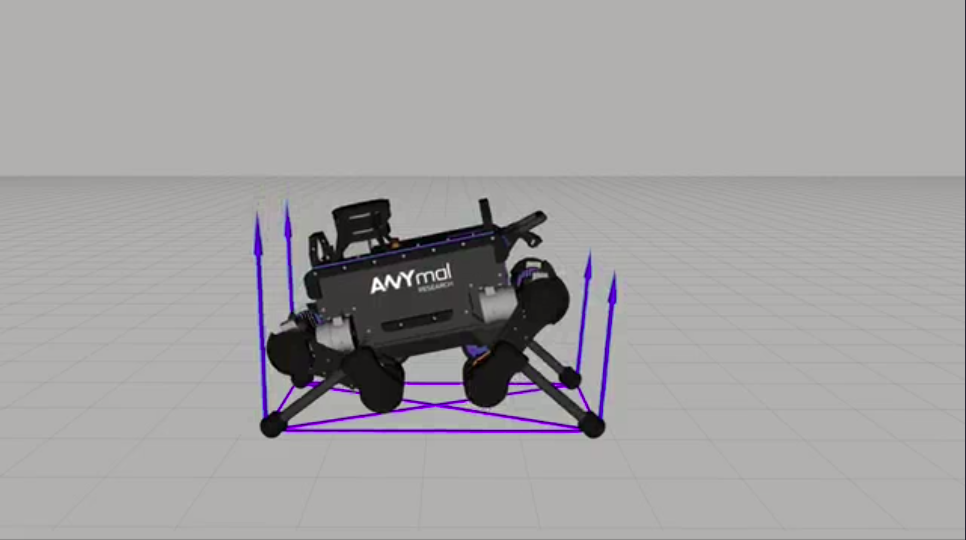}
\end{subfigure}\hfil

\medskip
\begin{subfigure}{0.155\textwidth}
  \includegraphics[width=1.02\linewidth]{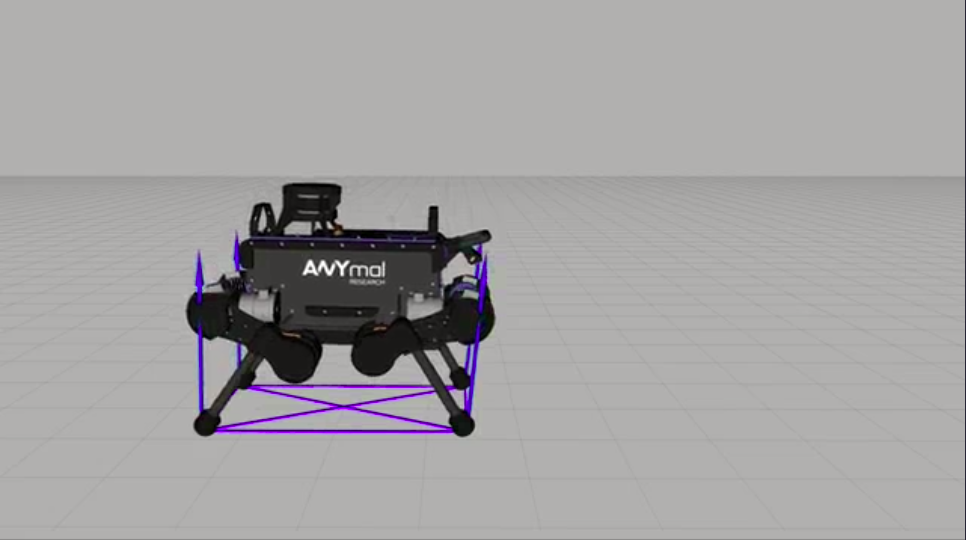}
\end{subfigure} 
\begin{subfigure}{0.155\textwidth}
  \includegraphics[width=1.02\linewidth]{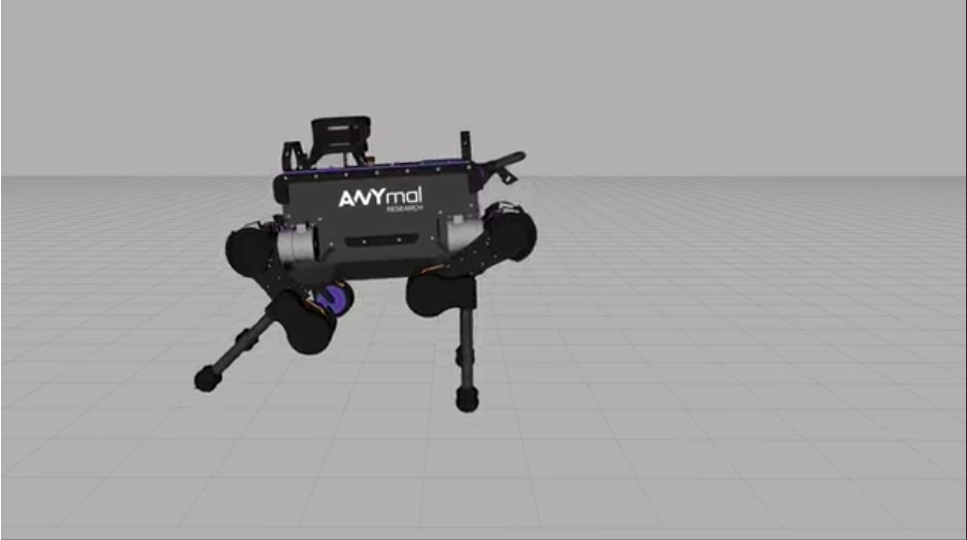}
\end{subfigure} \hfil
\begin{subfigure}{0.155\textwidth}
  \includegraphics[width=1.02\linewidth]{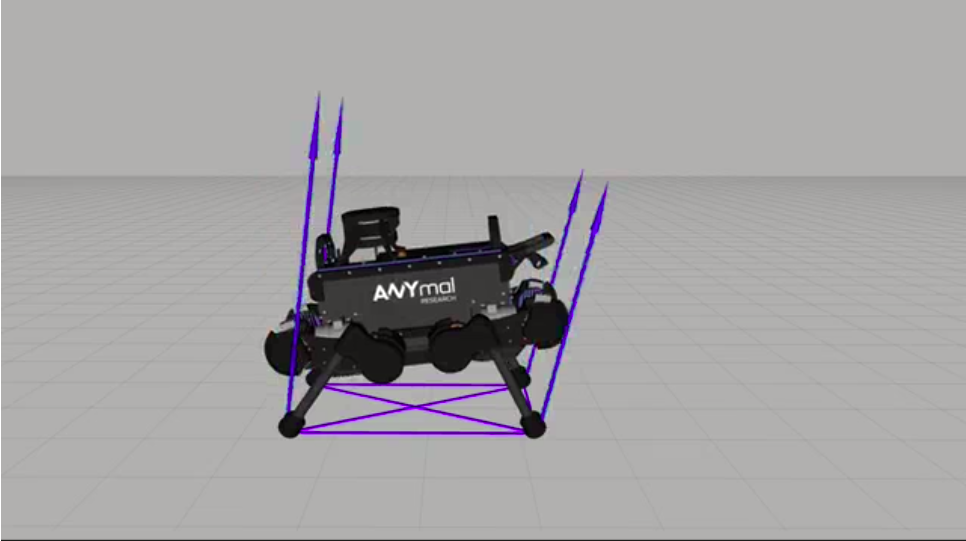}
\end{subfigure} \hfil
\caption{Snapshots of the ANYmal robot performing a forward jump of 30~cm with inertia shaping and without inertia shaping. First row: forward jump with inertia shaping. Second row: forward jump without inertia shaping.}
\label{fig:jumpForward_snapshot_anymal}
\vspace{-5mm}
\end{figure}

In the trajectory optimization, we assign a large weight to the \textit{final state target} and \textit{energy consumption}. For ANYmal, a forward jump motion of 30~cm is generated with LL-SRBM and SRBM, as show in Fig. \ref{fig:jumpForward_snapshot_anymal}. It can be seen that for trajectories generated with LL-SRBM model, the feet stretch out more in the air than trajectories generated with SRBM. This is because the optimizer tries to increase the inertia around $y$ axis to reduce the change of base pitch angle. The stretched feet in the flight phase also help to prevent early touchdown of the robot.

\begin{figure}[t]
    \centering 
\begin{subfigure}{0.2\textwidth}
  \includegraphics[width=0.9\linewidth, height=3.5cm]{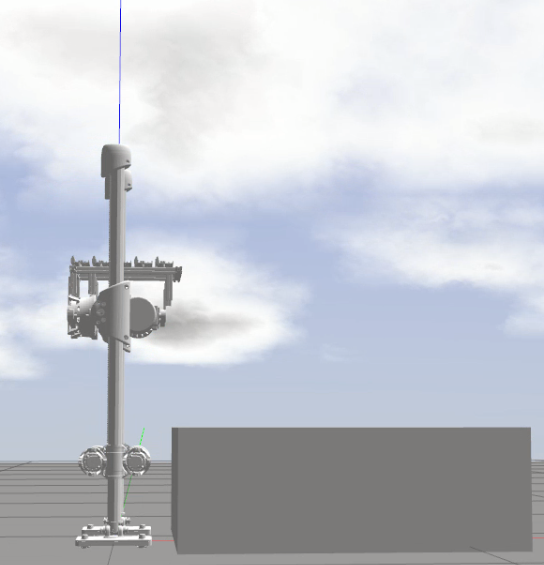} 
  \caption{}
\end{subfigure} 
\begin{subfigure}{0.2\textwidth}
  \includegraphics[width=0.9\linewidth,height=3.5cm]{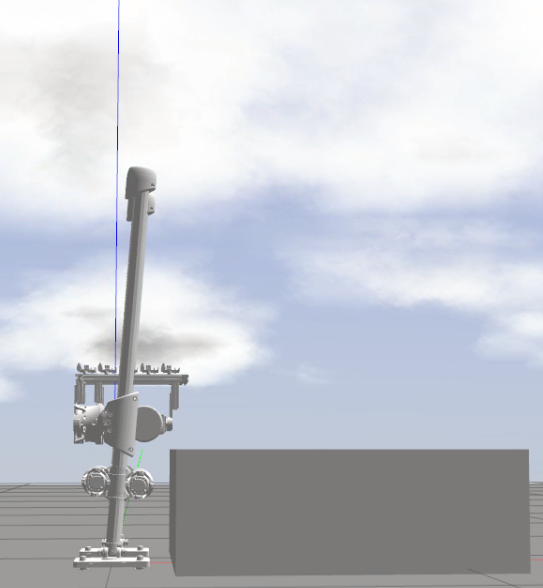}
  \caption{}
\end{subfigure}\hfil

\medskip
\begin{subfigure}{0.2\textwidth}
  \includegraphics[width=0.9\linewidth, height=3.5cm]{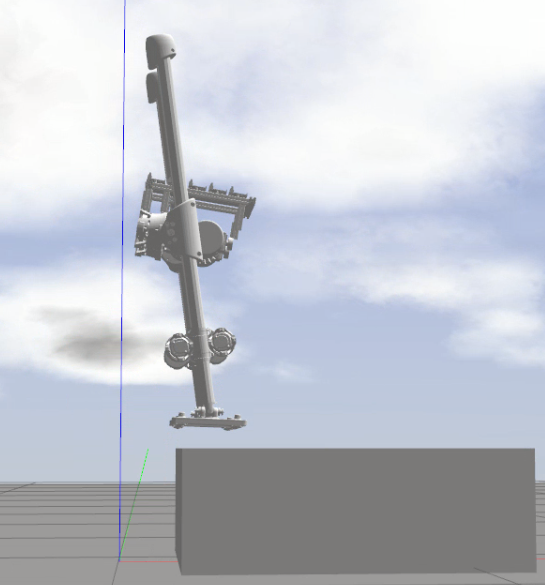}
  \caption{}
\end{subfigure} 
\begin{subfigure}{0.2\textwidth}
  \includegraphics[width=0.9\linewidth, height=3.5cm]{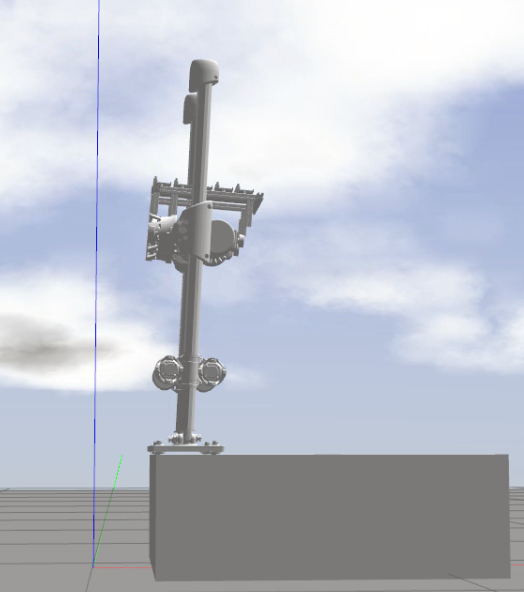}
  \caption{}
\end{subfigure} \hfil
\caption{Snapshots of SLIDER performing a forward jump onto a 35~cm box using LL-SRBM. $\text{(a)}\sim\text{(b)}$: takeoff phase, (c): flight phase, (d): landing phase.}
  \label{fig:jumpUp_snapshot_SLIDER}
\vspace{-5mm}
\end{figure}

We also tried jump onto a box for SLIDER with LL-SRBM. The maximum height of the box SLIDER can jump on is 35~cm, with a forward jump length of 20~cm, as shown in Fig. \ref{fig:jumpUp_snapshot_SLIDER}. Considering that SLIDER is 1.2~m high, the robot can jump onto a box equal to 30\% of its total height. Compared with the anthropomorphic robot design which has knees, SLIDER's straight legs have a larger range of motion, allowing SLIDER to reach the same jump height with a relatively smaller CoM height. Because there are no knee joints on legs of SLIDER, the feet can go all the way up until the ankles touch the pelvis. That is the reason why SLIDER can jump onto a high box. 

\subsection{Computation Time}
We recorded the computation times of trajectory optimizations for various dynamic jump motions. For the bipedal robot we used 60 knots and each knot has 59 variables (26 for states and 33 for control inputs). For the quadrupedal robot we used 50 knots and each knots has 53 variables (26 for states and 27 for control inputs). Although the complexity of LL-SRBM has increased compared with SRBM, the computation time with LL-SRBM only increase by an average of 33.6\% with respect to that with SRBM, as shown in Table \ref{tab:table_1}.
\begin{table}[h]
\caption{Solve Times for Planners.}
\label{table_2}
\begin{center}
\setlength\tabcolsep{1.5pt}
\begin{tabular}{|c|c|c|c|c|}
\hline
  &\thead{quadruped \\ forward jump}  & \thead{quadruped \\twist jump} & \thead{biped \\ forward jump} & \thead{biped \\twist jump} \\
\hline
\textbf{SRBM} & 1.50s & 2.45s & 6.33s & 5.44s\\
\hline
\textbf{LL-SRBM} & 1.53s & 2.56s & 9.54s & 7.37s\\
\hline
\end{tabular}
\end{center}
\vspace{-5mm}
\label{tab:table_1}
\end{table}

\section{CONCLUSIONS}
This paper proposes an unified model with inertia shaping for planning highly dynamic jumps of legged robots. This model allows the motion planner to improve the jumping performance by actively changing the centroidal inertia. In the meanwhile, this paper also proposes a novel contact detection method using the norm of average spatial velocity. The twisting jump and forward jump experiments on bipedal robot SLIDER and quadrupedal robot ANYmal show the improved jump performance after using the proposed model and the robustness of the controller to unforeseen impacts. In the future, we are interested in speeding up the computation to re-plan the jump motion with the proposed LL-SRBM.



\section*{APPENDIX}

	\section*{Quaternion to Rotation Matrix Conversion}\label{appendix1}
	Given a quaternion $\boldsymbol{q} =q_w+q_x \boldsymbol{i}+q_y \boldsymbol{j}+q_z \boldsymbol{k}$ which represents the orientation of the single rigid body in world frame, the equivalent rotation matrix representation can be derived as:
	\begin{equation*}
	\boldsymbol{R}=
	\left[\begin{array}{ccc}
							  1-2\left(q_y^{2}+q_z^{2}\right) & 2 q_x q_y-2 q_w q_z & 2 q_w q_y+2 q_x q_z \\
							  2 q_x q_y+2 q_w q_z & 1-2\left(q_x^{2}+q_z^{2}\right) & 2 q_y q_z-2 q_w q_x \\
							  2 q_x q_z-2 q_w q_y & 2 q_w q_x+2 q_y q_z & 1-2\left(q_x^{2}+q_y^{2}\right)
	\end{array}\right]
	\end{equation*}

	\section*{Derivative of Quaternion}
	Followed by \cite{graf2008quaternions}, given a rigid body with quaternion $\boldsymbol{q} =q_w+q_x \boldsymbol{i}+q_y \boldsymbol{j}+q_z \boldsymbol{k}$ and with angular velocity $\bm{\omega}$, the derivative of the quaternion $\dot{\boldsymbol{q}}$ can be calculated as:

	$$\dot{\boldsymbol{q}}=\frac{1}{2}  \boldsymbol{q} \circ {\bm{\omega}}$$
	$$\left[\begin{array}{c}
				\dot{q}_{x} \\
				\dot{q}_{y} \\
				\dot{q}_{z} \\
				\dot{q}_{w}
	\end{array}\right]
	= \frac{1}{2} \left[\begin{array}{c}
												  {q}_{x} \\
												  {q}_{y} \\
												  {q}_{z} \\
												  {q}_{w}
	\end{array}\right]
	\circ
	\left[\begin{array}{c}
	{\omega}_{x} \\
	{\omega}_{y} \\
	{\omega}_{z} \\
			  	0
	\end{array}\right]
	$$

	$$\left[\begin{array}{c}
				\dot{q}_{x} \\
				\dot{q}_{y} \\
				\dot{q}_{z} \\
				\dot{q}_{w}
	\end{array}\right]=\frac{1}{2}\left[\begin{array}{cccc}
												  q_{w} & -q_{z} & q_{y} & q_{x} \\
											q_{z} & q_{w} & -q_{x} & q_{y} \\
											-q_{y} & q_{x} & q_{w} & q_{z} \\
											-q_{x} & -q_{y} & -q_{z} & q_{w}
	\end{array}\right]\left[\begin{array}{l}
								w_{x} \\
								w_{y} \\
								w_{z} \\
								0
	\end{array}\right]$$
	




\section*{ACKNOWLEDGMENT}
 This work is supported by the CSC Imperial Scholarship, EPSRC UK RAI Hubs NCNR (EP/R02572X/1), FAIR-SPACE(EP/R026092/1). The authors would like to thank Digby Chappell for helpful discussions.

\bibliographystyle{IEEEtran}
\bibliography{bibliography.bib}

\end{document}